\documentclass[sigconf]{acmart}

\AtBeginDocument{%
  }

\copyrightyear{2023}
\acmYear{2023}
\setcopyright{acmlicensed}\acmConference[CIKM '23]{Proceedings of the 32nd ACM International Conference on Information and Knowledge Management}{October 21--25, 2023}{Birmingham, United Kingdom}
\acmBooktitle{Proceedings of the 32nd ACM International Conference on Information and Knowledge Management (CIKM '23), October 21--25, 2023, Birmingham, United Kingdom}
\acmPrice{15.00}
\acmDOI{10.1145/3583780.3615514}
\acmISBN{979-8-4007-0124-5/23/10}

\usepackage{subfig, tikz}
\usepackage{booktabs}
\usepackage{balance} 

\definecolor{darkblue}{RGB}{25, 25, 112}

\begin{document}

\title{Can Knowledge Graphs Simplify Text?}

\author{Anthony Colas*}
\email{acolas1@ufl.edu}
\affiliation{%
  \institution{The University of Florida}
  \city{Gainesville}
  \state{Florida}
  \country{USA}
}

\author{Haodi Ma*}
\email{ma.haodi@ufl.edu}
\affiliation{%
  \institution{The University of Florida}
  \city{Gainesville}
  \state{Florida}
  \country{USA}
}

\author{Xuanli He}
\email{xuanli.he@ucl.ac.uk}
\affiliation{%
  \institution{University College London}
  \city{London}
  \country{United Kingdom}
}

\author{Yang Bai}
\email{baiyang94@ufl.edu}
\affiliation{%
  \institution{The University of Florida}
  \city{Gainesville}
  \state{Florida}
  \country{USA}
}

\author{Daisy Zhe Wang}
\email{daisyw@cise.ufl.edu}
\affiliation{%
  \institution{The University of Florida}
  \city{Gainesville}
  \state{Florida}
  \country{USA}
}

\renewcommand{\shortauthors}{Anthony Colas, Haodi Ma, Xuanli He, Yang Bai, \& Daisy Zhe Wang}


\begin{abstract}
 Knowledge Graph (KG)-to-Text Generation has seen recent improvements in generating fluent and informative sentences which describe a given KG. As KGs are widespread across multiple domains and contain important entity-relation information, and as text simplification aims to reduce the complexity of a text while preserving the meaning of the original text, we propose \textbf{KGSimple}, a novel approach to unsupervised text simplification which infuses KG-established techniques in order to construct a simplified KG path and generate a concise text which preserves the original input's meaning. Through an iterative and sampling KG-first approach, our model is capable of simplifying text when starting from a KG by learning to keep important information while harnessing KG-to-text generation to output fluent and descriptive sentences. We evaluate various settings of the \textbf{KGSimple} model on currently-available KG-to-text datasets, demonstrating its effectiveness compared to unsupervised text simplification models which start with a given complex text. Our code is available on \href{https://github.com/acolas1/KGSimple}{\color{darkblue} \texttt{GitHub}}\footnotemark.

\end{abstract}



\maketitle

\newcommand\blfootnote[1]{%
\begingroup
\renewcommand\thefootnote{}\footnote{#1}%
\addtocounter{footnote}{-1}%
\endgroup
}
\blfootnote{* These authors contributed equally.}
\footnotetext[1]{\small \href{https://github.com/acolas1/KGSimple}{{\color{darkblue}\texttt{https://github.com/acolas1/KGSimple}}}}

\section{Introduction}
Text simplification (TS) is characterized as a text revision task, with the constraint that the output text should be easier to read than the input text. The primary objective of TS is to propagate information to a more expansive audience, including individuals with lower literacy levels~\cite{stajner-2021-automatic}, those with reading disabilities~\cite{chen2017automatic}, non-native speakers~\cite{nisioi-etal-2017-exploring}, and individuals lacking specialized knowledge within specific domains, such as medically related documents~\cite{abrahamsson2014medical, van2019evaluating}. Therefore it can also enhance various natural language processing (NLP) tasks that necessitate less complex texts for optimal results, including question answering~\cite{dadu2021text}, information extraction~\cite{van-etal-2021-may-help}, and machine translation~\cite{vstajner2016can}.

Models employing generative supervised learning for TS have typically adhered to a sequence-to-sequence framework, whereby a complex body of the text is translated into a simplified sentence in an autoregressive fashion~\cite{xu-etal-2016-optimizing, nisioi-etal-2017-exploring}. Conversely, unsupervised models typically depend on parsing and restructuring sentences to generate text that is both simpler and semantically relevant~\cite{dong-etal-2019-editnts, kumar-etal-2020-iterative}. The generative models frequently surpass their counterparts that depend on editing mechanisms and rules for sentence simplification, although they necessitate the availability of abundant parallel sentence pairs for effective training, which may not be consistently accessible. Moreover, while supervised models execute the translation of complex sentences through a series of continuous steps, unsupervised TS has predominantly concentrated on employing edit-based approaches. These approaches involve rule-driven operations, including sentence reordering, splitting, and deletion, providing a higher level of interpretability, but are inherently limited by their predefined rule sets.

Recent investigations into unsupervised text simplification (TS) have delved into the domain of search-based methodologies. These methods utilize predefined scoring functions that prioritize factors such as simplicity, fluency, and meaning retention in a sentence as per the studies conducted by ~\citet{laban-etal-2021-keep, vo2022unsupervised, dehghan-etal-2022-grs}. Similar to supervised counterparts, these methodologies can also incorporate a generation phase to maintain the integrity of sentence structure. Despite these advancements, current unsupervised strategies are limited in their capacity to generate highly fluent sentences. 
This limitation arises due to the constraints imposed by edit operations on the structural and syntactical elements of the input sentences.

\begin{figure*}
\centering
\includegraphics[width=\textwidth]{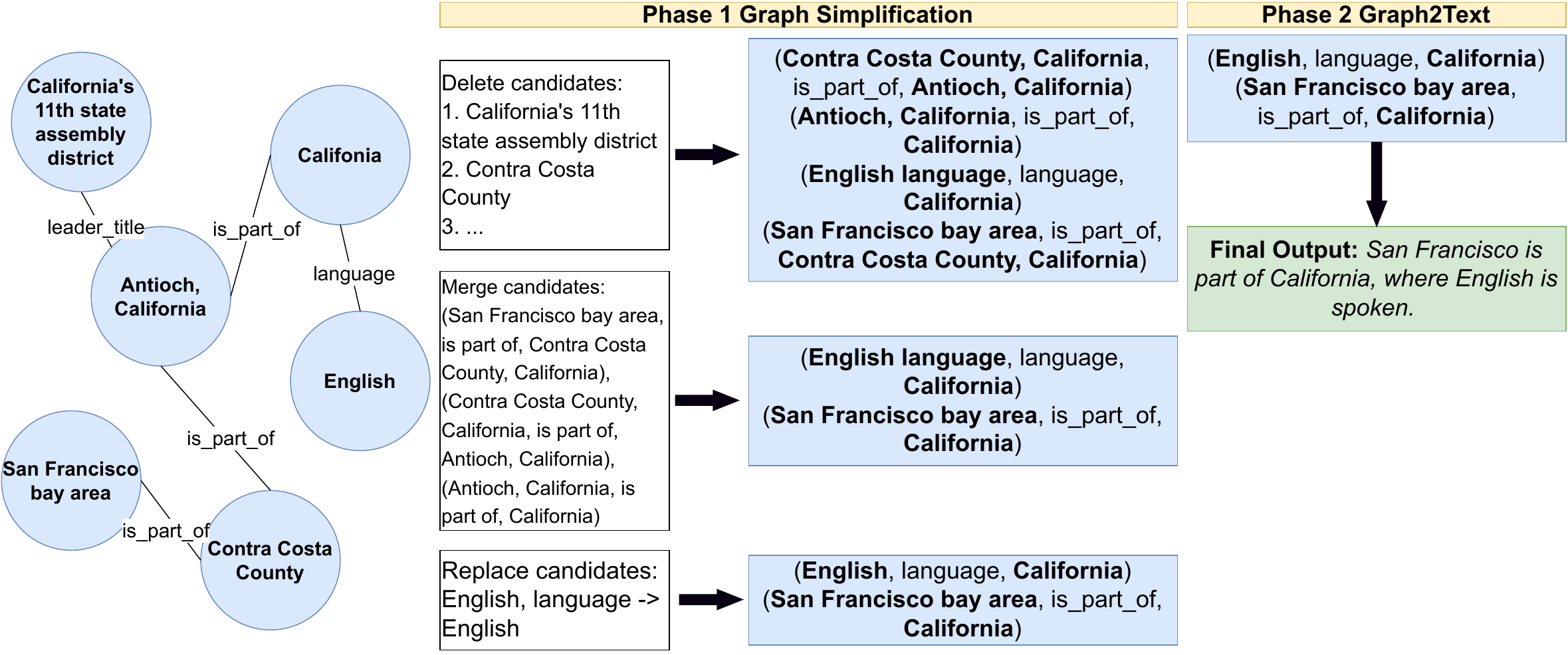}
\caption{Illustration of the KGSimple framework: from input KG to generated text.}
\label{fig:motivating_example}
\end{figure*}


As there is a vast amount of data stored in knowledge graphs (KGs), including patient records~\cite{rotmensch2017learning}, legal procedures~\cite{filtz2017building}, and event-related information~\cite{colas2021eventnarrative} it is important to convey such information to a wide audience. KG-to-text models have seen recent improvements in transcribing such structured data into fluent natural language sentences which preserve the meaning of the KG~\cite{ke-etal-2021-jointgt, colas-etal-2022-gap}.
Still, because of the vast amounts of data stored in KGs, some of the data may be irrelevant and difficult to understand when transcribing to a user. Furthermore, as the data stored in KGs can act as a compact representation of narratives, the data stored in KGs can be used to help interpret such TS models, especially those that leverage a text generation step. Thus, we propose an unsupervised framework that infuses TS with KG-to-text generation to generate simplified texts that preserve the meaning of the original input KG while generating more fluent texts than existing unsupervised TS approaches. Whereas current TS approaches rely on input texts, we extend TS to the KG domain. 
Our framework, which we call KGSimple, is a sampling-based approach that utilizes graph operations to simplify the input's contents while leveraging pre-trained KG-to-text generative models to produce fluent and relevant text with respect to the input KG. An example is illustrated in Figure~\ref{fig:motivating_example}. We adopt an iterative graph sampling to produce simplified KGs akin to Markov Chain Monte Carlo (MCMC) sampling, where at each step a local operation is performed on a KG to reduce the structural and syntactic complexity of the graph.
We then generate a text from the proposed graph, either accepting or rejecting the current KG-text pair according to a heuristically defined scoring function. We explore multiple sampling algorithms to produce a simplified text, including simulated annealing (SA)~\cite{van1987simulated} to search for a simplified KG.

Our contributions are as follows:
\begin{itemize}
    \item We propose KGSimple, an unsupervised KG-text simplification framework that can produce simplified text based on an input KG. Note, that our algorithm can produce both sentence and paragraph-level simplification.
    \item We devise up to three sampling techniques to simply more complex KGs and integrate existing KG-to-text models on the simplified KGs to generate coherent and simplified texts.
    \item We experiment on existing KG-to-text data and compare it to existing text-centric unsupervised TS systems, demonstrating that KGSimple can at times outperform these models, specifically in generating fluent simplified text in an interpretable fashion.  
\end{itemize}

\section{Related Work}
\subsection{Text Simplification}

A variety of generative supervised models have been employed, including sequence-to-sequence models~\cite{nisioi-etal-2017-exploring}, reinforcement learning techniques~\cite{zhang-lapata-2017-sentence}, and transformer architectures~\cite{vaswani2017attention}. These approaches have utilized external paraphrase databases~\cite{zhao-etal-2018-integrating}, complexity-weighted loss~\cite{kriz-etal-2019-complexity}, syntactic rules~\cite{maddela-etal-2021-controllable}, and complexity features found within the text~\cite{martin-etal-2020-controllable} to improve their performance.

On the other hand, edit-based supervised models have been developed to simplify complex sentences by leveraging parallel complex-simple sentences. \citet{alva-manchego-etal-2017-learning} have proposed a method that learns the keep, replace, and delete operations, while \citet{dong-etal-2019-editnts} have developed an end-to-end generative model to learn where to apply edit operations. Recent work on supervised edit-based text simplification has focused on predicting token-level edit-operations in a non-autoregressive fashion~\cite{omelianchuk-etal-2021-text} or editing the complex text through a fixed pipeline~\cite{agrawal-etal-2021-non}. However, unlike KGSimple, these approaches require large amounts of parallel supervised data and focus mainly on sentence and word-level edits. In contrast, our KG-level edits can produce changes in the output at various granularities.

Several semi-supervised methods have been proposed for text simplification. For example, ~\citet{zhao2020semi} introduced a back-translation framework, while ~\citet{surya-etal-2019-unsupervised} employed a style-transfer technique, and ~\citet{martin-etal-2022-muss} fine-tuned BART approach. However, these methods still require non-aligned pairs of sentences, which limits their interpretability and controllability since they are based on generative networks. On the other hand, unsupervised edit-based approaches have been explored, where a pipeline of operations is applied to complex sentences~\cite{narayan-gardent-2016-unsupervised}.

Recently, iterative approaches have been proposed, where text simplification is modeled as a search problem ~\cite{kumar-etal-2020-iterative}. These methods have integrated text generation~\cite{laban-etal-2021-keep, dehghan-etal-2022-grs} and back-translation~\cite{vo2022unsupervised} as a step in their framework. Although KGSimple is also an iterative revision-based approach, it differs from these methods in that it leverages the knowledge graph's condensed storage of information as a starting point instead of plain text. Moreover, while previous work on text simplification has mostly focused on greedy selection processes, KGSimple considers various sampling strategies, inspired by adjacent text generation tasks\cite{li2020unsupervised}.

\subsection{KG-Text Generation}
Previous research on models that convert KG into natural language text first employed GNNs to encode the neighborhood structure of the graph before decoding the information into text~\cite{velivckovicgraph, koncel-kedziorski-etal-2019-text, he-etal-2020-scene}. Recent studies in the field of KG-to-text have explored the efficacy of pre-trained language models~\cite{ribeiro-etal-2021-investigating} and have developed pre-training tasks to acquire knowledge of the KG representation~\cite{hoyle-etal-2021-promoting}. These models have also incorporated graph-based features into transformers, such as a node's relative position~\cite{schmitt-etal-2021-modeling} and the graph's topology~\cite{ke-etal-2021-jointgt, colas-etal-2022-gap}, in order to enhance KG encoding. While the previously mentioned research has concentrated on narrating the KG, we are the first to apply such KG-to-text models to unsupervised text generation, specifically, text simplification.

\section{Approach}
\begin{figure}[t!]
\centering
\includegraphics[width=0.45\textwidth]{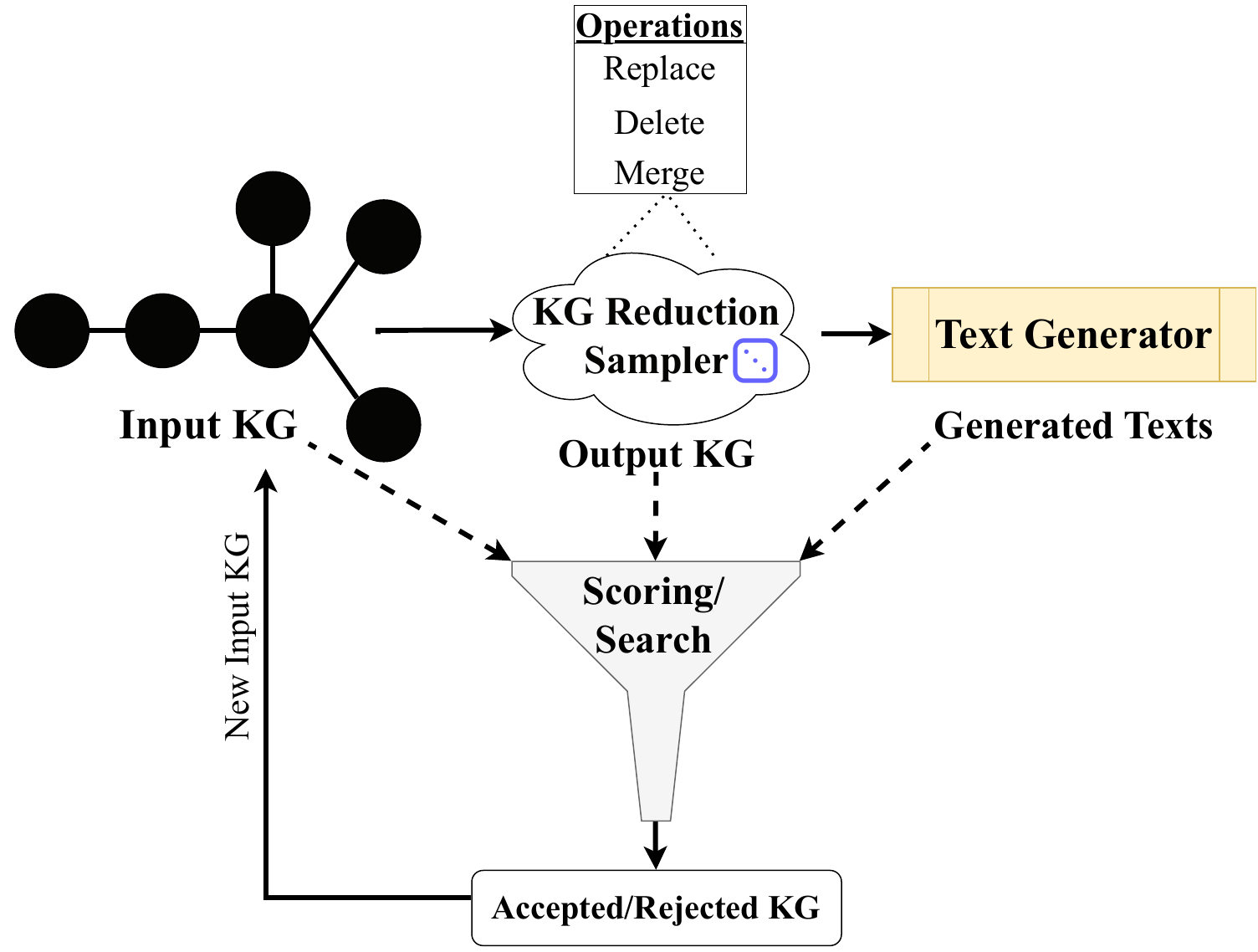}
\caption{The KGSimple framework: 1) the KG Reduction Sampler produces a simplified KG, 2) the text generator (GAP) produces a text representation, and 3) the scoring function evaluates the KG-text pair according to a given condition.}
\label{fig:kgsimple_overview}
\end{figure}
\subsection{Overview}
The proposed framework, KGSimple, comprises a two-fold approach that employs an iterative process to optimize a given reward function. As depicted in Figure~\ref{fig:kgsimple_overview}, given an input KG $g$, the framework performs the following steps: (1) sample a simplified graph $g'$ through diverse graph reduction operations on the structure and syntactic content of the original graph (refer to Section~\ref{sec:operations_subsection}), and (2) generate a corresponding text $y_{g'}$ via a KG-to-text generator (refer to Section~\ref{sec:generator_subsection}). Both $g$ and $y_{g'}$ are then evaluated based on a pre-defined scoring function (refer to Section~\ref{sec:scoring_subsection}).

At each iteration of the framework, starting with KG $g_0$, a reduction operation is applied to the KG, and a pre-trained generative model produces a text that corresponds to the current KG. Our scoring function considers both soft and hard constraints on the KG-Text pair and determines whether to retain or discard the proposed KG-Text pair for use in the next iteration. If the KG-Text pair is rejected, KGSimple will randomly sample another operation on the KG to generate a new text from the generator.

Our scoring function is heuristically defined, and therefore, to ensure the credibility of our results, we employ three distinct conditional acceptance criteria outlined in Section~\ref{sec:scoring_subsection}. First, we accept scores that are greater than or equal to zero. Second, we consider scores greater than the previous score to indicate progress. Finally, we utilize simulated annealing (SA) as a technique to encourage the model to explore beyond local minima. By employing these acceptance criteria, we aim to ensure the robustness and accuracy of our scoring system.

\subsection{Operations}
\label{sec:operations_subsection}
We detail the operation set sampled by KGSimple that simplifies a KG at each step. We propose candidates by sampling one of the operations that can modify entities/relation tokens and reduce the structure of the original KG. 

\vspace{2mm}
\textbf{Delete. }
In this operation, we propose a new candidate KG by eliminating one of the entities in the original KG. To ensure the preservation of crucial information after each deletion, we utilize a TF-IDF score for each phrase, which is calculated using the English Wikipedia. At each iteration, the model first identifies the entity node(s) with the lowest degree or least amount of connected edges. Subsequently, if there are multiple candidates, we exploit the TF-IDF score to delete the node with the least frequency and the least centrality. The delete operation enables us to eliminate redundant branches for graph-level simplification.

\textbf{Replace. }
To semantically simplify the graph, at each iteration, we select the most complex and least frequent word based on the IDF score. After selecting the complex word, we employ a two-stage approach to generate feasible replacements: (1) we obtain candidate phrases by leveraging a complex word dictionary from Simple PPDB~\cite{pavlick-callison-burch-2016-simple}, and (2) we consider a candidate phrase to be an appropriate substitution when it has a lower IDF score than the complex word.

For instance, suppose that the original KG contains the phrase: \textit{(`The menu' `offer' `vegetarian alternatives')}. In that case, our model can replace the word \textit{`alternatives'} with \textit{`options'}. Similarly, for another phrase such as \textit{(`Annual Avant Garde Festival of New York', `inception', `1963')}, we replace the word \textit{`inception'} with \textit{`beginning'}. These operations significantly enhance the semantic simplicity of the graphs.

\textbf{Merge. }
To further simplify the graphs, we consider merging operations to certain types of edges:
\begin{itemize}
    \item[1)] Transitive edges: For two edges $(e_1, r_1, e_2)$ and $(e_2, r_2, e_3)$, our model merges them into $(e_1, r_1|sep|r_2, e_3)$. \\ For example, with \textit{(Steven Gerrard, play, Liverpool F.C.)} and \textit{(Liverpool F.C., locate in,  the county of Merseyside)}, our model will merge them into \textit{(Steven Gerrard, play|sep|locate in,  the county of Merseyside)}
    \item[2)] Edge pairs that share both head and tail entities: For two edges $(e_1, r_1, e_2)$ and $(e_1, r_2, e_2)$, our model merges them into $(e_1, r_1|sep|r_2, e_2)$. \\ For example, \textit{(`2013–14 Albanian Superliga', `location', `Albania')} and \textit{(`2013–14 Albanian Superliga', `country', `Albania')}. 
    \item[3)] Edge pairs that share the same head or tail entity: For two edges $(e_1, r_1, e_2)$ and $(e_1, r_2, e_3)$ or $(e_2, r_1, e_1)$ and $(e_3, r_2, e_1)$, our model merges them into $(e_2, r_1|sep|r_2, e_3)$.
\end{itemize}

Furthermore, we use the sum of the IDF of all the phrases in the two edges to select the most complex and least informative pair to merge. The Merge operation improves our model's ability to further simplify the graph structure.

\subsection{Generator}
\label{sec:generator_subsection}
As part of the second phase of KGSimple, the proposed knowledge graph $g_t$ is subjected to a generative KG-to-text model in each iteration, which produces a natural language output text $y$. This output text is generated based on the following probability formula:
$$P(y_{g_t}) = \prod_{i=1}^{n} P(w_i | w_1^{i-1},g_t)$$,
where $w_i$ is the token generated at position $i$, $n$ is the output length, and $y_{g_t}$ is the output text corresponding to the proposed graph at iteration $t$.

By leveraging text generative models like GAP~\cite{colas-etal-2022-gap}, our framework produces coherent and relevant sentences that describe the contents of the given graph. Note that multiple passes through this model, coupled with our scoring function, help improve the output of these models iteratively. This iterative process can reduce hallucination, which refers to the model generating content during the decoding step that lacks support from the given input data.

\subsection{Scoring}
\label{sec:scoring_subsection}
Several previous studies have examined unsupervised TS, using heuristic-based reward functions to evaluate various syntactic and semantic aspects of the output text $y$ based on individual metrics~\cite{hinton2002training, kumar-etal-2020-iterative, laban-etal-2021-keep}. These reward functions aim to ensure that the output text is fluent ($S_{fl}$), preserves the original meaning ($S_{mp}$), and is simpler in content than the input ($S_{sim}$), while also rejecting poor quality candidates generated through the iterative process. To refine these reward functions for the KG case, we propose using the product of all individual rewards to ensure balance across all criteria. 

\subsubsection{Fluency}
To ensure that the generated candidates are grammatically correct and fluent, we use the syntactic log-odds ratio (SLOR) score~\cite{pauls-klein-2012-large}, which is defined as the sentence log-probability normalized by the unigram probability and sentence length, as the fluency score $S_{fl}$. The SLOR score works well as a proxy for fluency, as previous work has found that SLOR highly correlates with the human evaluation of grammatically acceptability~\cite{lau2017grammaticality} and fluency in the sentence compression task~\cite{kann-etal-2018-sentence}.

Given a unigram probability model, pre-trained language model (LM), and text, SLOR assigns a score to the text $y_t$ at iteration $t$ as:
$$SLOR(y_t) = \frac{1}{|y_t|} \left( \ln(p_{LM}(y_t)) - \ln(p_{U}(y_t)) \right),$$
where $|y_t|$ is the length of the text, $p_{LM}(y_t)$ is the probability of $y_t$ under a given LM, $p_{U}$ is the unigram probability measures as: 
$$p_{U}(y_t) = \prod_{w \in y_t} p(w)$$, 
where $p(w_i)$ is the unconditional probability of token $w$ produced by a unigram-text model.

SLOR penalizes an LM's probability by the unigram probability, stabilizing the LM probability in the presence of rare tokens as the LM score may assign a lower probability to those texts containing rare tokens. This ensures that rare tokens do not penalize a text's fluency score which is important when evaluating the output text of a KG containing rare entity tokens as we do not want to penalize these rare tokens. Where previous works have used recurrent neural networks to calculate the LM probability for SLOR, we use a pre-trained GPT-2~\cite{brown2020language} model with byte-pair encoding (BPE). To keep the score between [0,1] we avoid calculating the log of the model scores, instead opting to use their raw probabilities:
$$S_{fl}(y_t)=\frac{1}{|y_t|} \left(p_{LM}(y_t) - p_{U}(y_t) \right)$$.

\subsubsection{Salience}
Salience has been typically defined as the similarity between a candidate text and pseudo-reference text, where a text with high salience has similar semantic meaning and both contain particularly important words~\cite{erkan2004lexrank}. To measure the salience of a generated simplified text, we use the F1 of BERTScore~\cite{zhangbertscore} between the generated text at the current iteration $y_t$ and the text generated from the original KG $y_0$, defined as:
\begin{align*}
S_{mp}(y_0, y_t) &= \text{BERTScore}(y_0,y_t)
\end{align*}

 The BERTScore leverages the pre-trained BERT model when calculating the cosine similarity between two texts. Where previous work~\cite{kumar-etal-2020-iterative, dehghan-etal-2022-grs} uses the cosine similarity as a hard filter, our salience calculation acts as a soft score which is multiplied by our other scoring metrics, as the BERTScore typically falls between a very narrow high range in multiple settings~\cite{hanna-bojar-2021-fine}.


\subsubsection{Simplicity}
Following previous work, we evaluate the simplicity $S_{simp}$ of a generated text $y$ with the Flesch Reading Ease (FRE) score~\cite{kincaid1975derivation}, which measures the readability of a text based on the number of total sentences, words, and syllables.
A lower score indicates that a text is more difficult to read.
While the FRE score typically ranges from [0,100], the highest possible score is 121.22 with no theoretical lower bound. Therefore, to bound the range from [0, 1] we set the simplicity score to:
$$S_{si}(y_t) = \frac{FRE(y)- \lambda}{121.22-\lambda},$$ where $\lambda$ is a tunable hyper-parameter representing the lowest possible raw FRE score in a given dataset. For our purposes, we set $\lambda = -30$, in our experiments as a lower bound.

\subsubsection{Entity Score}
Unlike prior work in unsupervised TS, our task is a KG-to-text simplification task. Therefore, we check whether the generated text $y_t$ accurately narrates the currently proposed KG $g_t$. To do so, we penalize the reward function if the KG-to-text model hallucinates any new entities that do not appear in the generated text. Because a KG is a compact and structured representation, it may be the case that the corresponding text would need to mention new entities in order to produce fluent sentences. We, therefore, impose a soft constraint if the model hallucinates new entities. The entity score $S_{en}$ is defined as:
$$S_{en}(y_t,g_t) = 1- \frac{|E^y_t \setminus E^g_t|}{|E^g_t|},$$
where $E^g_t$ and $E^y_t$ are the entities in $g_t$ and $y_t$.

\subsubsection{Hard Constraints}
In order to prevent our model from producing invalid results, we define hard (binary) constraints on the proposed simplified KG and output text which cause the score to zero out if triggered.

As our sampled operations induce a reduction of the KG, our first hard constraint is a graph brevity penalty score $S_{gb}$, which is set to zero if the number of triples in proposed KG $g_t$ is reduced below a certain threshold. In our experiments, we set the threshold as the ratio $r_{op} = 0.6 \geq \frac{|T^g_{t-1}|}{|T^g_t|} $, where $|T^g_{t-1}|$ and $|T^g_t|$ denote the number of triples in the previous and proposed graph, respectively.

Our second phase generates text from KG's, which may hallucinate entities that are likely to appear in a generated narration. Since our approach aims to remove unimportant entities' information, we verify that those entities that were specifically deleted from $g$ do not appear in the simplified text. If these entities do appear in the generated text, the deleted entity score $S_{de}$ is set to zero, automatically rejecting the sample.

\subsubsection{Overall}
As the various scores capture distinct facets of the simplified input, we compute the overall reward function as:
\begin{equation*}\label{eq:overall_reward}
  \begin{aligned}
S(y_0, y_t,g_{t-1},g_t) &= S_{fl}(y_t)\cdot S_{mp}(y_0, y_t)\\
  &\cdot S_{si}(y_t) \cdot S_{en}(y_t,g_t)\\
  &\cdot S_{gb}(g_{t-1}, g_t) \cdot  S_{de}(y_t,g_0)
  \end{aligned}
 \end{equation*}

\begin{table*}[!ht]
    \caption{Comparison of unsupervised text simplification models on WebNLG and DART. Len, CR, SC, and SR, refer to the Length, Compression Ratio, Syllable Count, and Syllable Ratio. EO, Add, and Delete, refer to Entity Text Overlap, Entity Text Added, and Entity Text Deleted. The constituency tree Height and Diameter are denoted by CTH and CTD, and the fluency (CoLA) and salience (BERTScore) are denoted by F and S. GM is the geometric mean between the CR, SR, F, and S scores.  $\uparrow$/$\downarrow$ indicates that a higher/lower value is better.}
    \label{tab:webnlg-simp-results}
    \subfloat[Result on WebNLG\label{tab:result:1a}]{
        \begin{tabular}{@{}l|cccccc|ccc|ccc@{}}
            \toprule
             & Len $\downarrow$  & CR $\downarrow$   & SC $\downarrow$   & SR $\downarrow$ &    CTH $\downarrow$  & CTD $\downarrow$ & EO   & Add $\downarrow$  & Delete $\uparrow$ & F $\uparrow$ & S $\uparrow$ & GM $\uparrow$ \\ \midrule
            Baseline (GAP)  & 37.45 & 1.00  & 47.63 & 1.00  & 23.59 & 36.28 & -     & -     & -     & 0.62  & 1.00 & - \\
            
            EditTS          & 33.17 & 0.90  & 40.82 & 0.89  & 22.74 & 33.73 & 0.89  & 0.29  & 0.53  & 0.48  & \textbf{0.96} & 0.27 \\
            
            USDP            & 24.78 & 0.69  & \textbf{27.53} & \textbf{0.60} & 20.92 & \textbf{26.55} & 0.59  & \textbf{0.06}  & 1.26 & 0.54 & 0.93 & 0.50\\ \midrule
            
            GRS             & 26.04 & 0.71  & 31.87 & 0.68   & \textbf{20.31} & 29.51    & 0.75  & 0.68  & 1.39  & 0.56  & 0.93 & 0.47\\ \midrule
            
            Prev            & 31.39 & 0.85  & 38.70 & 0.82  & 22.22 & 32.51 & 0.75  & 0.53  & 1.26  & 0.63  & 0.93 & 0.35\\
            
            SA              & 23.69 & 0.65  & 28.84 & 0.63  & 20.79 & 27.89 & 0.61  & 0.62  & 1.91  & 0.62  & 0.90 & 0.52\\
            
            Zero-best       & 30.01 & 0.82  & 36.94 & 0.79  & 22.33 & 31.90 & 0.75  & 0.56  & 1.34  & \textbf{0.64} & 0.92 & 0.39\\
            
            Zero-last       & \textbf{23.07} &\textbf{ 0.64} & 28.12 & 0.61 & 20.58 & 27.19 & 0.59 & 0.59 & \textbf{1.91} & 0.61 & 0.90 & \textbf{0.53} \\ \bottomrule
            \end{tabular}%
    }
    \hfill
    \subfloat[Result on DART\label{tab:result:1b}]{
        \begin{tabular}{@{}l|cccccc|ccc|ccc@{}}
        \toprule
         & Len $\downarrow$  & CR $\downarrow$   & SC $\downarrow$   & SR $\downarrow$ &    CTH $\downarrow$  & CTD $\downarrow$ & EO   & Add $\downarrow$  & Delete $\uparrow$ & F $\uparrow$ & S $\uparrow$ &  GM $\uparrow$\\ \midrule
        
        Baseline (T5)   & 28.94 & 1.00  & 36.75 & 1.00  & 19.61 & 30.66 & -    & -    & -    & 0.51 & 1.00 & -\\
        
        EditTS          & 26.94 & 0.94  & 33.03 & 0.91  & 19.48 & 29.63 & 0.67 & 0.18 & 0.25 & 0.43 & 0.96 & 0.22 \\
        
        USDP            & 18.21 & 0.67  & \textbf{20.77} & \textbf{0.59}  & 18.06 & \textbf{21.03} & 0.30 & \textbf{0.05} & 1.13 & 0.37 & 0.93 & 0.46 \\ \midrule
        
        GRS             & 21.78 & 0.75  & 27.25 & 0.75  & 17.69 & 25.91 & 0.53 & 0.45 & 0.90 & 0.51 & 0.94 & 0.42 \\ \midrule
        
        Prev            & 26.58 & 0.93  & 33.02 & 0.90  & 18.85 & 29.03 & 0.58 & 0.16 & 0.46 & 0.51 & \textbf{0.96} & 0.24 \\
        
        SA              & 19.52 & 0.71  & 23.55 & 0.67  & 17.28 & 24.78 & 0.46 & 0.30 & 1.03 & 0.50 & 0.92 & 0.46 \\
        
        Zero-best       & 24.20 & 0.86  & 29.46 & 0.82  & 18.63 & 27.79 & 0.52 & 0.25 & 0.76 & \textbf{0.52} & 0.94 & 0.33 \\
        
        Zero-last       & \textbf{17.72} & \textbf{0.64}  & 21.49 & 0.61  & \textbf{16.81} & 22.75 & 0.40 & 0.37 & \textbf{1.20} & 0.51 & 0.91 & \textbf{0.51}\\ \bottomrule
        \end{tabular}
    }
    
\end{table*}

\subsection{Search Algorithms}
\label{sec:search_algo}
In this paper, we investigate various acceptance conditions for the reward scoring function previously defined in our framework. These acceptance conditions include accepting the current proposal if and only if its score is greater than zero, accepting the proposal if its score is higher than the previous score, and employing an SA approach that promotes optimal solution exploration in the model's initial stages. The SA approach sets a threshold that is controlled by probability. If the candidate KG $g_t$ is accepted at iteration $t$, it is processed next. However, if it is rejected, the algorithm proposes another operation on $g_{t-1}$.

While previous work has applied SA to other text generation tasks~\cite{li2020unsupervised}, we remodel SA for TS. In the SA approach, at each iteration $t$ a KG proposal $g_t$ is accepted with probability:
$$p(A|g_{t-1},g_t,y_{t-1},y_t, T) = \min\{1, e^{-\Delta E/T}\}$$
$$\Delta E = S(y_{t},g_0,g_{t}) - S(y_{t-1},g_0,g_{t-1}),$$
where $T$ is the temperature or tolerance, initially set to a high value and decreased via a cooling rate~\cite{kirkpatrick1983optimization}.

\section{Experiments}
\subsection{Data}
To evaluate the KGSimple framework, we use the WebNLG~\cite{gardent-etal-2017-creating} KG-to-text dataset and DART~\cite{nan-etal-2021-dart} structured data record to text generation dataset, as the KGs contained within these datasets were constructed for natural language text generation. WebNLG is a hand-crafted triple-to-text dataset that contains graphs from DBPedia~\cite{auer2007dbpedia} with up to 7 triples paired with reference texts. DART is a human-annotated and automatic-converted dataset incorporated from WikiSQL~\cite{zhong2017seq2sql}, WikiTableQuestion~\cite{pasupat2015compositional}, WebNLG 2017, and Cleaned E2E~\cite{novikova2017e2e} with 21.6 words on average for each table. 

We extract the more complex KGs from the WebNLG and DART datasets, considering those KGs containing more than 3 triples as complex. We thus experimented on 583 samples on WebNLG and 500 samples on DART. As there currently only exists unsupervised text-based approaches for TS, for a fair comparison we first generate a text with a KG-to-text model from each of the KGs. We use GAP~\cite{colas-etal-2022-gap} for WebNLG and T5~\cite{ribeiro-etal-2021-investigating} for DART to generate such golden sentences. The generated text then serves as the complex text input in each of the baselines.

\subsection{Competing Models}
We evaluate KGSimple under the three aforementioned search criteria, namely: 1) greater than previous, 2) greater than zero, and the SA algorithm. As in previous work, we consider the complex input as an upper bound for each of the evaluation metrics. Each text generated by GAP/T5 acts as the golden reference, as it is the starting point used in all of the evaluated approaches.

For the unsupervised competing approaches, we compare them with state-of-the-art edit-based iterative approaches. First, we compare against~\cite{kumar-etal-2020-iterative}, an iterative revision-based approach composed of a delete (RM), extraction (EX), lexical substitution (LS), and reordering (RO) operation. We include the RM+EX+LS+RO setting and denote it as EditTS. Next, we compare against USDP~\cite{vo2022unsupervised}, a sentence simplification system that uses a dependency tree structure to decode a structurally simpler output. The generated output is then back-translated to English to generate lexical simplifications. Finally, we compare to GRS~\cite{dehghan-etal-2022-grs}, an iterative framework that uses explicit edit operations to reduce complex text, including a generative paraphrase module. In our experiments, we use the PA+DL setting of GRS which performs both deletion and paraphrasing on an input text. 

\subsection{KGSimple Settings}
For the KG-to-text generator, we use the GAP w/ type encoding~\cite{colas-etal-2022-gap} and T5~\cite{nan-etal-2021-dart} models, which were fine-tuned on simple graphs (1-3 triples) from the WebNLG and DART dataset, respectively. GAP is a KG-to-text model that combines graph-aware elements into the BART~\cite{lewis-etal-2020-bart} text generative model. We leave all of the default hyper-parameters for GAP and T5, and set the beam size to 5, with a repetition penalty of 1.0. T5 is a state-of-art data-to-text generation model and has been evaluated on DART. We use T5-base in our experiment and follow the same setting as in \cite{ribeiro-etal-2021-investigating}. 

We set the probability of sampling the $\left[ \text{delete}, \; \text{replace}, \; \text{merge} \right]$ operations to a uniform distribution of $[\frac{1}{3}\text{, }\frac{1}{3}\text{, }\frac{1}{3}]$. We run each input for 50 iterations with a batch size of 64. All components of our scorer are weighted equally, where we calculate the sentence probability for the fluency scorer using GPT-2 and calculate the BERTScore using the HuggingFace evaluate library~\footnote{https://github.com/huggingface/evaluate}. For the \textit{greater than previous} and \textit{SA} search algorithms, we normalize all scores between 0 and 1. To extract entities from the output text in the entity scorer, we use the spaCy~\cite{ines_montani_2023_7715077} named entity recognition module. All experiments were performed on 2 NVIDIA A-100 GPUs. 

\subsection{Metrics}
To evaluate all of the models, we measure intrinsic metrics associated with text simplicity, including the text length, number of syllables, semantic tree height, and semantic treewidth. As we propose a KG simplification approach, we also measure on entity-based metrics. These include counting the entity overlap between the input/output (entity text overlap), how many entities were added to the output (entity text added), and how many entities were removed from the input (entity text deleted). As we want to ensure the output is coherent, we evaluate the \textit{fluency} and \textit{similarity} using the CoLA~\cite{warstadt2019neural} and BERTScore. We use CoLA, as the LM perplexity does not account for repetition or grammar when scoring. For CoLA, we report the probability that a text is indeed acceptable. We exclude the FKGL from our evaluation, as FKGL was designed for a text of at least 200 words~\cite{xu2015problems, alva-manchego-etal-2020-asset}. Additionally, ~\citet{tanprasert-kauchak-2021-flesch} have recently found that FKGL can be easily manipulated via basic post-processing steps and argue to instead report the individual components of FKGL, i.e., sentence length and syllable count. Therefore, we report these, along with coherence metrics (fluency and saliency), and a geometric mean which combines the size metrics with the coherence metrics. Unlike prior work on unsupervised TS, where the data still comes from a supervised dataset, our datasets do not contain any simplified text. We, therefore, exclude calculating the BLEU~\cite{papineni-etal-2002-bleu} or SARI~\cite{xu2016optimizing} scores.

\section{Results}
We present the results on WebNLG and DART in table~\ref{tab:webnlg-simp-results}.  We refer to the complex generated text data as \textit{Baseline}. For the KGSimple search algorithm with the "greater than zero" condition, we report both the last accepted iteration (\textit{Zero-last}) and the best-scored iteration (\textit{Zero-best}), as this condition is more loosely constrained than our other conditions. We denote the "greater than prev" and SA algorithms as \textit{Prev} and \textit{SA}, respectively.
\subsection{WebNLG}
From table~\ref{tab:result:graph_web}, we can observe that KGSimple with SA outperforms the KGSimple configuration with the \textit{Prev} search constraint in terms of intrinsic simplification metrics, with a text compression ratio of 0.65, syllable ratio of 0.63, and constituency tree height/diameter of 20.79 and 27.89. Although \textit{Zero-last} produces slightly shorter sentences than the \textit{SA} approach, the \textit{SA} approach produces more fluent sentences. Conversely, the \textit{Zero-best} approach and the more tightly conditioned \textit{Prev} method produce more fluent sentences. Hence, for the unsupervised text simplification approaches there may be some trade-off between length and fluency which future work can further study. When scoring the geometric mean between the length ratios, fluency, and saliency, we find that the \textit{SA} and \textit{Zero-last} search constraints perform similarly, where \textit{Zero-last} produces slightly shorter texts and \textit{SA} produces slightly more fluent texts. In terms of text entity metrics, the \textit{SA} and \textit{Zero} approaches perform similarly, where \textit{Prev} keeps more of the entities from the original text. All KG-based simplification approaches perform consistently and similarly for fluency and saliency, where they seem to be able to produce grammatically correct and semantically similar texts. This may be in part because the generator excels at producing coherent text from KGs, while our simplification approach manages to keep and combine the important KG features.

Compared to the text-based unsupervised models, KGSimple can generate shorter texts, while producing similar results in terms of syllable counts and constituency tree height/diameter. KGSimple also produces more fluent texts compared to GRS and USDP. While the model proposed by~\citet{kumar-etal-2020-iterative} has similar fluency scores compared to KGSimple, their text lengths do not decrease by much, with a compression ratio of 0.85, compared to the KGSimple \textit{SA} compression ration of 0.65 and \textit{Zero-last} compression ratio of 0.64. For the entity-based metrics, KGSimple deletes more entities than the text-based approach, but can also introduce some newer entities. When compared to USDP and GRS, KGSimple has similar text entity overlap. The KG-based approaches all outperform the text-based approach in linguistic acceptability (fluency). Note that although GRS uses the CoLA as a constraint in their search algorithm, KGSimple still outperforms GRS on this metric. While GRS is the only other generative-based unsupervised approach, KGSimple consistently outperforms the model. Therefore, KGSimple appears to be a better overall system when desiring more variability and flexibility in simplifying sentences. The text-based approaches do achieve a slightly higher BERTScore than the KG-based approaches. However, this may be because the edit text-based approaches focus on directly modifying (cutting) the original text, while our KG-based approach needs to produce a new text from the currently proposed KG at each iteration.

\begin{table}
  \begin{minipage}[t]{0.45\linewidth}
    \centering
    \caption{KGSimple output graph analysis: number of triples (Triples), in/out triple ratio (TR), entity ratio (ER), and relation ratio (RR).}
    \label{tab:webnlg-graph-simp}
    \subfloat[Result on WebNLG\label{tab:result:graph_web}]%
    {\begin{tabular}{@{}lcccc@{}}
    \toprule
    & Triples & TR & ER & RR \\ \midrule
Prev      & 3.31 & 0.67 & 0.84 & 0.74 \\
SA   & 2.09 & 0.45 & 0.66 & 0.50 \\
Zero & 2.03 & 0.44 & 0.65 & 0.49 \\
\bottomrule
    \end{tabular}
    }
    
    \subfloat[Result on DART\label{tab:result:graph_dart}]%
       {\begin{tabular}{@{}lcccc@{}}
    \toprule
    & Triples & TR & ER & RR \\ \\ \midrule
Prev      & 4.32 & 0.78 & 0.87 & 0.83 \\
SA   & 2.19 & 0.41 & 0.61 & 0.45 \\
Zero & 2.03 & 0.39 & 0.58 & 0.42 \\
\bottomrule
    \end{tabular}
    }
  \end{minipage}
  \hfill
  \begin{minipage}[t]{0.45\linewidth}
    \centering
   \caption{Overall ratio analysis for the accepted individual individual \textit{Delete (D)}, \textit{Replace (R)}, and \textit{Merge (M)} graph operations for KGSimple.}
   \label{tab:operation-simpleWebNLG}
   \subfloat[Result on WebNLG\label{tab:result:op_web}]%
    {\begin{tabular}{@{}lccc@{}}
    \toprule
    & D & R & M  \\ \midrule
Prev & 0.33   & 0.34    & 0.33  \\
SA   & 0.26   & 0.45    & 0.28  \\
Zero & 0.26   & 0.47    & 0.28  \\ \bottomrule
    \end{tabular}
    }
    
    \subfloat[Result on DART\label{tab:result:op_dart}]%
       {\begin{tabular}{@{}lccc@{}}
    \toprule
    & D & R & M  \\ \midrule
Prev & 0.55   & 0.34    & 0.11  \\
SA   & 0.50   & 0.33    & 0.17  \\
Zero & 0.47   & 0.35    & 0.18  \\ \bottomrule
    \end{tabular}
    }
  \end{minipage}
\end{table}

\subsection{DART}
From table~\ref{tab:result:1b}, compared to the results from WebNLG, we can observe a consistent performance of KGSimple on DART. With SA, KGSimple achieves better performance on the intrinsic simplification metrics (length and syllable count), compared to \textit{Prev}. On the other hand, the method with a tighter condition, \textit{Zero-best} compared to \textit{Zero-last}, again generated more fluent sentences. This provides further evidence of the possible tradeoff between the length and fluency of text-simplification methods. Generally, all KG-based methods achieve similar performance for fluency and saliency, while \textit{Prev} preserves more entities from the original text as in WebNLG. Again here, as with WebNLG, the \textit{Zero-last} and \textit{SA }approaches gave the best geometric means between compression ratio, syllable ratio, fluency, and saliency scores.

On DART, unlike in the experiments on WebNLG, USDP generates shorter texts than those generated by the KGSimple \textit{SA} algorithm. However, note the discrepancy in the fluency score, where USDP has the lowest linguistically acceptable generated texts. So while the texts may be the shortest, they may not be as useful as those generated by KGSimple, which may have a slightly better balance between the metrics. While EditTS and GRS achieve compatibly similar fluency performance against KGSimple, KGSimple \textit{Zero-last} and \textit{SA} generate shorter texts with lower compression ratios of 0.64 and 0.71. On the entity-based evaluation, KGSimple adds fewer new entities while deleting more entities and less entity overlap than all text-based approaches except USDP. This may be because USDP directly removes sentences/clauses from the text. On linguistic metrics, KGSimple, on the other hand, generates more fluent texts with the highest fluency score of 0.52 and the highest saliency score of 0.96. Additionally, compared to the generative-based approach GRS, our \textit{SA} and \textit{Zero-last} KGSimple approach generates shorter sentences on average, while maintaining a consistent text fluency score.

\begin{table*}[ht]
\centering
\small
\begin{tabular}{@{}p{7cm}p{7cm}@{}}
\toprule
  CT &
  ST \\ \midrule
  the bronze ataturk monument (izmir) is located in izmir and was inaugurated on 27 july 1932. the capital of the country is ankara and the largest city istanbul. the country's leader is ahmet davutoğlu and the currency is the turkish lira. &
  the bronze atatürk monument is located in the largest city of izmir and was inaugurated on the 27th of july, 1932. \\ \\
  
  the american submarine nr-1 has a 3.8m ship beam and a top speed of 8.334 km/h. it was built by the company general dynamics electric boat and was launched on january 25, 1969. &
  the american submarine nr-1 has a top speed of 8.334 km/h and was launched on january 25, 1969. \\ \\

  the city of aarhus, which has to its northeast mols, is governed by a magistrate. the city is the location of the school of business and social sciences at the aarhus university which is affiliated with the european university associations which has its headquarters in brussels &
  the school of business and social sciences at the aarhus university has its headquarters in the city of aarhus and has strong connections with the academic community.
  \\\\

  the city of lahore is served by the allama iqbal international airport which is located in pakistan. the airport is operated by the pakistan civil aviation authority and has a runway length of 2900.0. &
the city of lahore is served by the allama iqbal international airport which is operated by the civil aviation authority.\\\\
  
  the 14th new jersey volunteer infantry monument is located in the district of the monocacy national battlefield, frederick county, maryland, united states. the monument was established on 11th july 1907 and belongs to the category of historic districts in the united states. &
  the 14th new jersey volunteer infantry monument is located in frederick county, maryland, united states. \\ \bottomrule
\end{tabular}%
\caption{Examples from the  WebNLG and DART datasets. Left: complex text (CT). Right: simplified text (ST) results generated by the KGSimple framework.}
\label{tab:kgsimple-examples}
\end{table*}
Generally, the performance of KGSimple on DART is consistent with the result on WebNLG when compared to the text-based baselines. Compared to the text-based baselines which rely on using separate models for text generation and simplification, KGSimple integrates the KG-to-text models in its simplification framework, to produce syntactically simpler and more fluent texts, while keeping fewer entities from the original sentences and introduces fewer entities for text generation.

\section{Analysis}

\subsection{Graphs}
As KGSimple is a graph-centric approach, we compare the input KGs to the reduced KGs from which the simplified text is generated in table~\ref{tab:webnlg-graph-simp}. 
We compare the graph statistics in each of our search settings, including the number of triples (Triples), output-to-input KG size ratio (TR), output-to-input unique entity ratio (ER), and output-to-input unique relation ratio (RR). From the table, we can see that although the simulated annealing (SA) algorithm imposed a tighter constraint than the \textit{Zero} algorithm, which accepts the current proposal as long as the overall score was greater than zero, the \textit{SA} was able to reduce the KG almost as close to the \textit{Zero-last} condition. In contrast, as expected, the \textit{Prev} algorithm has the overall largest number of final triples and restricts the graph reduction to 67\% and 78\% of the original KG. Thus, our input-output graph analysis confirms that the \textit{SA} condition can encourage further KG reduction (search space exploration) for KG-text simplification, while the more restrictive \textit{Prev} condition may get stuck at local minima. This is reflected when analyzing both the entity (ER) and relation (RR) analysis, where the \textit{Prev} condition contains a larger ratio compared to \textit{SA} and \textit{Zero}.

\subsection{Operations}
The KGSimple framework is an iterative approach that proposes changes to a KG based on sampled graph-reduction operations. Table~\ref{tab:operation-simpleWebNLG} compares the relative acceptance ratios of the delete, replace, and merge operations. From the results on WebNLG, in all search algorithms, our most accepted operation was \textit{replace}, while \textit{delete} and \textit{merge} were not accepted as often. For DART, the most accepted operation was \textit{delete}, while \textit{merge} was not as frequently accepted as in the WebNLG dataset. We currently use node centrality and TD-IDF to find those nodes to delete from the KG. While these nodes may not be as important in the KG, they may at times convey information that is crucial when transcribing the text either for semantic or fluency purposes. The discrepancy in both datasets for \textit{merge} may be because the datasets contain different ontologies, graph structures, and relations. WebNLG may contain more sets of triples that are acceptable to merge, while DART's KGs may contain a more sporadic structure. The \textit{merge} operation may be further modified depending on the overall structure of the KG's and ontology found in the specific dataset. 

\begin{table}[ht]
    \caption{Case study from the  WebNLG. Show the original text, simplified result by USDP, and result by KGSimple framework step by step}
    \label{tab:case-study}
    \centering
    \small
    \begin{tabular}{@{}p{7cm}}        
        \toprule
        \begin{tabular}{@{\hspace{2em}}c@{\hspace{2em}}}
          \scshape Original Text: \\
          \midrule
        \end{tabular} \\
        The alternative name for the AMC Matador is Vam Classic and it is classed as a mid-size car. It is made in Kenosha, Wisconsin, and has an AMC straight 6 engine.
        \\ \midrule
        \begin{tabular}{@{\hspace{2em}}c@{\hspace{2em}}}
          \scshape USDP: \\
          \midrule
        \end{tabular} \\
        and it is classed - <mask> as a mid - size car .
        \\ \midrule
        \begin{tabular}{@{\hspace{2em}}c@{\hspace{2em}}}
          \scshape KGSimple: \\
          \midrule
        \end{tabular} \\
        \textbf{Graph 0: }(Vam classic, \textbf{alternative name}, AMC Matador), (Kenosha, Wisconsin, assembly, AMC Matador), (mid-size car, class, AMC Matador), (AMC straight-6 engine, engine, AMC Matador) \\
        \vspace{-\baselineskip}
        \begin{center}
            $\downarrow$
        \end{center}
        \vspace{-\baselineskip} \\
        \textbf{Graph 1: }(Kenosha, Wisconsin, \textbf{assembly}, AMC Matador), (mid-size car, class, AMC Matador), (AMC straight-6 engine, engine, AMC Matador)  \\
       \vspace{-\baselineskip}
        \begin{center}
            $\downarrow$
        \end{center}
        \vspace{-\baselineskip} \\
        \textbf{Graph 2: }(Kenosha, Wisconsin, found, AMC Matador), (mid-size car, class, AMC Matador), (\textbf{AMC straight-6 engine}, engine, AMC Matador) \\
        \vspace{-\baselineskip}
        \begin{center}
            $\downarrow$
        \end{center}
        \vspace{-\baselineskip} \\
        \textbf{Graph 3: }(Kenosha, Wisconsin, found, AMC Matador), (mid-size car, class, AMC Matador)\\
       \vspace{-\baselineskip}
        \begin{center}
            $\downarrow$
        \end{center}
        \vspace{-\baselineskip} \\
        \textbf{Simplified text: }The AMC Matador is classed as a mid-size car and is found in Kenosha, Wisconsin.
        \\ \bottomrule
    \end{tabular}%
\end{table}
\subsection{Examples}
We showcase five examples produced by KGSimple in Table~\ref{tab:kgsimple-examples} by comparing the text produced by the KG-to-text model on the original complex KG to the last accepted output text which was generated from the simplified KG. From these examples, we see that KGSimple has the capacity to simplify text by implicitly cutting sentences/clauses, replacing tokens, and merging phrases via the KG reduction operations. For example in the second example, where the text is about the "submarine nr-1", our approach removes phrases regarding the "ship beam" and "general dynamics electric boat", which were nodes in the complex KG. KGSimple also merges the two sentences via the merge operation on the KG. However, as our method is a generative approach (KG-to-text), one limitation is that the approach is prone to hallucination, similar to other generative approaches, such as GRS~\cite{dehghan-etal-2022-grs}. For example, in the first example referring to the "bronze ataturk monument", the simplified text implies that the monument is located in the largest city of "izimir", but the original text describes "istanbal" as the largest city. Second, replacement is limited by the dictionary contained in PPDB, as some complex words are not substituted. Nevertheless, as seen from all the examples, our approach can robustly cut out sentences, clauses, and specific details (entities/relations) which are not as important according to KGSimple's reduction step. For instance, in the last example about the "14th new jersey volunteer infantry monument", KGSimple simplifies some information about the exact location of the monument while fluently conveying its general location. Overall, the results across both the WebNLG and DART complex KGs show that KGSimple can coherently simplify the information found in the KGs and transform them into natural language sentences.

\subsection{Case Study}
To further demonstrate how KGSimple simplifies text by leveraging the graph compared to the text-based approaches, we show a detailed simplification in this section and compare it to the USDP model. As shown in Table~\ref{tab:case-study}, the original sentence is: \\
\textbf{The alternative name for the AMC Matador is Vam Classic and it is classed as a mid-size car. It is made in Kenosha, Wisconsin, and has an AMC straight 6 engine.}\\  Note, that as the original texts are all generated by the GAP KG-to-text model, there may be discrepancies between the graph and the text. For example, here, GAP replaced \textbf{assembly} in the graph with \textbf{made} when generating the original text. 
\\
We can see that the simplified text generated by USDP is shorter than KGSimple, but is less fluent and informative. In fact, USDP is more of a sentence-level method which usually keeps a sub-sentence as the simplified result. Compared to such an approach, KGSimple is able to simplify texts on various granularities, including the word level, by leveraging triple-level operations. Moreover, instead of selecting individual sentences, KGSimple also attempts to merge sentences together for further simplification via its merge operations on the KG. Following the operations also improves the interpretability of the simplification process.

Specifically, in step 1, KGSimple first identifies \textbf{alternative name} as an uncommon term and removes the related triple to simplify the graph. Then in step 2, KGSimple locates the term \textbf{assembly} as a complex term with replacement in the dictionary. Thus, it replaces the word with \textbf{found} to simplify the graph on the text aspect. Finally, in step 3, the model notices the term \textbf{AMC straight-6 engine} with the lowest TF-IDF score in the graph. Again, the delete operation behaves to remove this triple to further simplify the graph. As there is no further operation applied to the graph, GAP translates the graph back into the text as the final result. This example demonstrates how KGSimple simplifies texts on both the graph and text level, with explainable progress at each operation, which further shows the usefulness of KGs when attempting to interpret text simplification models.

\section{Conclusion}
We proposed KGSimple, a novel unsupervised KG-centric text simplification framework. We have demonstrated that text simplification can be accomplished via a KG-first approach, where the KG is reduced to its more crucial components and then narrated to natural language text via KG-to-text models. By considering both the KG and text in the reward, our approach has the potential to outperform text-centric unsupervised models as indicated by our results on the WebNLG and DART datasets, specifically able to generate less complex sentences while maintaining a relatively high fluency score. As we have developed the first framework for KG-text simplification, we encourage future work in other natural language generation tasks to explore KG-oriented approaches. We also encourage adapting our framework to other types of structured data such as tables or relational databases, where consumer healthcare data such as electronic medical records are often stored.
\section*{Acknowledgements}
This work is partially supported by the Arnold and Lisa Goldberg Endowed Professorship and DARPA under Award \#FA8750-18-2-0014 (AIDA/GAIA).

\bibliographystyle{ACM-Reference-Format}
\balance
\bibliography{sample-base,anthology}


\end{document}